\documentclass{article}
\usepackage{genai4health_2026}
\usepackage[T1]{fontenc}
\usepackage{amsmath,amssymb,amsthm,mathtools}
\usepackage{graphicx}
\usepackage{booktabs}
\usepackage{microtype}
\usepackage{xcolor}
\usepackage[colorlinks=true,allcolors=blue!55!black]{hyperref}
\setcitestyle{numbers,sort&compress}
\newtheorem{proposition}{Proposition}
\newcommand{\E}{\mathbb{E}}
\newcommand{\dd}{\mathrm{d}}
\newcommand{\cP}{\mathcal{P}}

\title{\textbf{Fairness Evaluation for Diffusion and Flow Models Should Account for the Generation Path: Implications for Clinical Imaging}}
\author{%
  Dharini Raghavan\textsuperscript{1},
  Amritpal Singh\textsuperscript{2},
  Anant Madabhushi\textsuperscript{1,2}\\[0.4em]
  \textsuperscript{1}Georgia Institute of Technology\\
  \textsuperscript{2}Emory University\\
  Corresponding author: \texttt{draghavan7@gatech.edu}
}
\date{}
\begin{document}
\maketitle
\begin{abstract}
Medical image generation models are often evaluated using images produced by a single sampling procedure. Diffusion and flow models can vary the number and timing of sampling steps or change numerical solvers without changing their trained parameters. These choices can alter generated images and the clinical measurements derived from them. We argue that fairness results obtained under one sampling procedure should not be assumed to hold under another, particularly when changes in sampling affect clinical task errors differently across patient groups. The sampling procedure is therefore part of the generative system being evaluated. We define subgroup differences in clinical task error across sampling procedures and provide a mathematical example showing that two groups can have identical outputs under the exact dynamics and a reference sampler but different errors under an alternative sampler. We also extend this result to stochastic generation under shared noise. An age-conditioned MRI synthesis experiment illustrates how changing the number and timing of sampling steps alters two image-derived measurements. We therefore argue that fairness evaluation should assess the sampling procedure used in practice and examine intermediate generation states when they help explain clinically relevant errors across patient groups. Analysis code is available at https://github.com/rdharini2001/generation-path-fairness
\end{abstract}


\section{Introduction}

Quantitative measurements derived from medical images are used to assess vascular disease, cancer, and other conditions, including in digital pathology~\cite{madabhushi2016pathology,bera2019digitalpath}. These measurements depend on accurate anatomical structures and tissue patterns, including vessels, lesion boundaries, and cell arrangements relevant to cancer studies~\cite{raghavanAbstract5078Sexspecific2026}. Small changes in these structures can affect clinical measurements even when the overall image appears realistic. Previous studies have shown that medical image generation and reconstruction methods can introduce false anatomical features or fail to preserve important structures. Conventional image similarity metrics may not adequately capture these errors~\cite{cohen2018distribution,antun2020instabilities,bhadra2021hallucinations,dohmen2025similarity,deo2025metrics}. This concern is particularly relevant for vascular structures, where small changes in vessel continuity or geometry may affect quantitative measurements~\cite{shit2021cldice,hu2019topology,stucki2023topologically}.

Generative models are being investigated for medical image synthesis, reconstruction, and quantitative analysis, including histopathology and MRI~\cite{aversa2023diffinfinite,luo2024aid}. In retinal imaging, for example, these methods have been explored for generating vascular images and studying microvascular measurements associated with disease~\cite{coronado2023synthetic,brownPhysicsinformedDeepGenerative2024,raghavan2026remodeling,raghavan2026cmdabstract,singhAIInformedRetinalBiomarkers2025}. Such applications require careful evaluation of whether generated images preserve the anatomical information needed for the intended clinical task. Related MRI and radiographic studies also rely on image-derived findings for clinical decisions and risk assessment~\cite{suma2022mriPetFusion,raghavan2023neuroimagingELM,singh2025spinaltb,singh2026cardiomorph,singh26a21554accOpportunisticHeart2026}. Visual similarity alone is insufficient when small structural differences can change measurements or downstream predictions.

For diffusion and flow-based models, the reliability of these measurements may also depend on the sampling procedure. These models generate images through successive numerical updates, and the number of sampling steps, their placement along the generation process, and the numerical solver can be changed without retraining the model~\cite{ho2020ddpm,song2021sde,lipman2023flow,liu2023rectified}. Such changes are often made to reduce computational cost or inference time. However, modifying the sampling procedure can change the generated images and their derived measurements, even when the model parameters remain fixed. A model evaluated using one sampling procedure may therefore behave differently when deployed with another, including in ways that affect clinical task errors across patient groups.

\textbf{We argue that fairness evaluation for diffusion and flow-based models should account for the sampling procedure used to generate the images and the clinical task for which those images are intended.} Fairness results obtained under one sampling procedure cannot be assumed to hold when that procedure changes. Evaluating intermediate generation states may also help explain when and how changes in sampling affect clinically relevant information. The relevant question is whether a change in sampling produces unequal changes in clinical task error across patient groups, after accounting for appropriate differences in anatomy, disease severity, and acquisition conditions.

We define subgroup differences in clinical task error across sampling procedures and construct an example in which group agreement under one sampler fails under another. An age-conditioned MRI synthesis experiment illustrates how sampling choices affect image-derived measurements. Together, these results motivate comparisons grounded in the intended clinical task and independent reference measurements.

\section{What Previous Work Tells Us}

\paragraph{Clinical image quality depends on the intended task.}
Medical image generation and reconstruction methods can produce realistic images while introducing false anatomical features or removing clinically important structures~\cite{cohen2018distribution,antun2020instabilities,bhadra2021hallucinations}. Studies of GAN-generated medical images and image-quality metrics have shown that visual similarity and distributional measures do not always reflect clinically relevant information~\cite{kelkar2023statistics,skandarani2023gans,dohmen2025similarity,deo2025metrics,koetzier2024synthetic,stein2023metrics,maierhein2024metrics}. For vascular imaging, measures of vessel continuity and topology provide more direct assessments of structural accuracy~\cite{gegundez2012evaluation,shit2021cldice,stucki2023topologically}. Work on clinical feedback for synthetic images and image quality in retinal prediction also shows the value of task-specific assessment~\cite{sun2023clinicalfeedback,raghavan2026quality}. In multiphase CT, PHORA evaluates LI-RADS predictions according to the distinction between ordered severity categories and clinically distinct special categories~\cite{raghavan2026phora}. Clinical outcomes also determine the appropriate measure in electrohysterogram-based preterm-birth prediction and cardiac-sound classification, while glucose-monitoring studies assess measurement errors by their clinical consequences~\cite{raghavan2023electrohysterogram,suma2024lightcardiacnet,suma2023nirglucose}. Outside medicine, AVID provides an example of task-specific evaluation by testing how target size, sensor misalignment, and image corruption affect feature-based comparisons in aerial imagery~\cite{raghavan2026avidweaklyalignedrgbinfrared}. These findings support evaluating generated images according to their intended clinical use, including the effect of changing the sampler on task-specific measurements.

\paragraph{Sampling choices can affect generated images.}
Diffusion and flow-based models generate images through successive transformations, including denoising, continuous flows, and stochastic processes~\cite{ho2020ddpm,song2021sde,lipman2023flow,liu2023rectified,albergo2025si,kim2024unsb}. Methods such as DDIM, DPM-Solver, and EDM, together with work on timestep schedules, parallel sampling, temporal attention, and distillation, have shown how generation can be made more efficient~\cite{song2021ddim,lu2022dpmsolver,karras2022elucidating,sabour2024align,shih2023parallel,liu2025temporal,zhou2024sfd}. These changes can affect generated images without modifying the trained model. Progressive distillation and consistency models provide further ways to accelerate generation, although these approaches also change the learned model~\cite{salimans2022progressive,song2023consistency}. The question for clinical use is whether the sampling procedure preserves the measurements on which subgroup comparisons depend.

\paragraph{Intermediate generation states contain useful information.}
Previous studies have shown that image features and learned representations change throughout diffusion-based generation~\cite{choi2022perception,luo2023hyperfeatures,biroli2024dynamical}. Li and Chen examined critical windows during which features emerge, including their relevance to fairness analysis~\cite{li2024windows}. Related work has investigated when features become detectable and how intermediate representations can be examined~\cite{li2025blink,alain2016probes,ganin2016dann}. These studies provide a basis for examining how generation dynamics relate to clinical measurements. For fairness evaluation, their value lies in explaining differences in task error across patient groups.

\paragraph{Fairness evaluation requires clinically meaningful comparisons.}
Previous research on generative model fairness has examined demographic representation, differences in image attributes, and methods for reducing bias~\cite{luccioni2023stablebias,friedrich2023fairdiffusion,shen2024finetune,ohayon2024perceptual,li2026fairgen}. Fairness measurement also requires defining which differences between groups are meaningful~\cite{truong2025valid}. Work on generative clinical language models provides another example: tests of deductive reasoning evaluate whether diagnostic constraints are followed rather than relying on common diagnostic patterns~\cite{raghavan2026deductive}. Data collection and synthetic-data sharing introduce related questions about population coverage, privacy, and fairness~\cite{aireadi2024,dombrowski2026generalize}. In medical imaging, disease severity, anatomy, acquisition conditions, and distribution shifts can influence subgroup performance~\cite{daneshjou2021transparency,chen2023algfair,schrouff2022fairshift,jin2024fairmedfm,yang2024fairgeneralization,viswanathan2024equitable}. Fairness evaluation must therefore examine clinically relevant errors while accounting for appropriate patient and acquisition factors. Observational subgroup comparisons should also be distinguished from causal claims~\cite{kusner2017counterfactual}. This concern is related to underspecification: models that perform similarly on standard benchmarks can behave differently in deployment~\cite{damour2022underspecification}.

Previous work establishes the importance of clinical image quality, numerical sampling, and intermediate generation states. Our position concerns the connection between them: a subgroup result obtained with one sampler must be examined again when the sampling procedure changes. The relevant comparison is clinical task error across samplers and patient groups, informed by intermediate states when they help explain the observed differences.

\section{Our position: evaluate the sampling procedure used in practice}

Let $G_{\theta,s}(U,Z)$ be the output of a trained generator given patient input $U$, random input $Z$, and sampling procedure $s$. The sampling procedure includes the solver, number and timing of steps, and precision. When a downstream method uses intermediate states, those states also form part of the system being evaluated. Let $T$ represent the clinical measurement or decision made from the output, and let $Y^*$ be an independent reference. For subgroup $A=g$ and relevant clinical or acquisition variables $Q=q$, we define the standardized task error as
\begin{equation}\label{eq:risk}
R_g(s)=\int \E\!\left[\ell\!\left(T(G_{\theta,s}(U,Z)),Y^*\right)\mid A=g,Q=q\right]w(q)\,\dd q,
\end{equation}
where $\ell$ is a prespecified loss and $w$ is a common distribution over values of $Q$ represented in both groups. The choice of adjustment variables matters. Adjusting for a factor that carries a disparity could hide it. The difference in task error between groups under sampler $s$ is $D(s)=R_b(s)-R_a(s)$. We compare this with the difference under a reference sampler $s_0$:
\begin{equation}\label{eq:change}
\Delta D(s,s_0)=\{R_b(s)-R_b(s_0)\}-\{R_a(s)-R_a(s_0)\}.
\end{equation}
Here, $D(s)$ measures the subgroup gap for a given sampler, while $\Delta D$ measures how that gap changes when the sampler changes. Their fairness interpretation depends on the error in each group, uncertainty estimates, clinically meaningful error margins, and relevant patient differences. For plausible sampling procedures $\cP$, we check whether subgroup errors and their differences remain acceptable across $\cP$.

Changing the sampler while holding the input and random noise fixed allows a controlled comparison of computational choices. Comparing patients by age or another characteristic is different: it is generally observational. This distinction separates a numerical intervention from an observational comparison between patient groups.

\begin{figure}[t]
\centering
\includegraphics[width=0.86\linewidth]{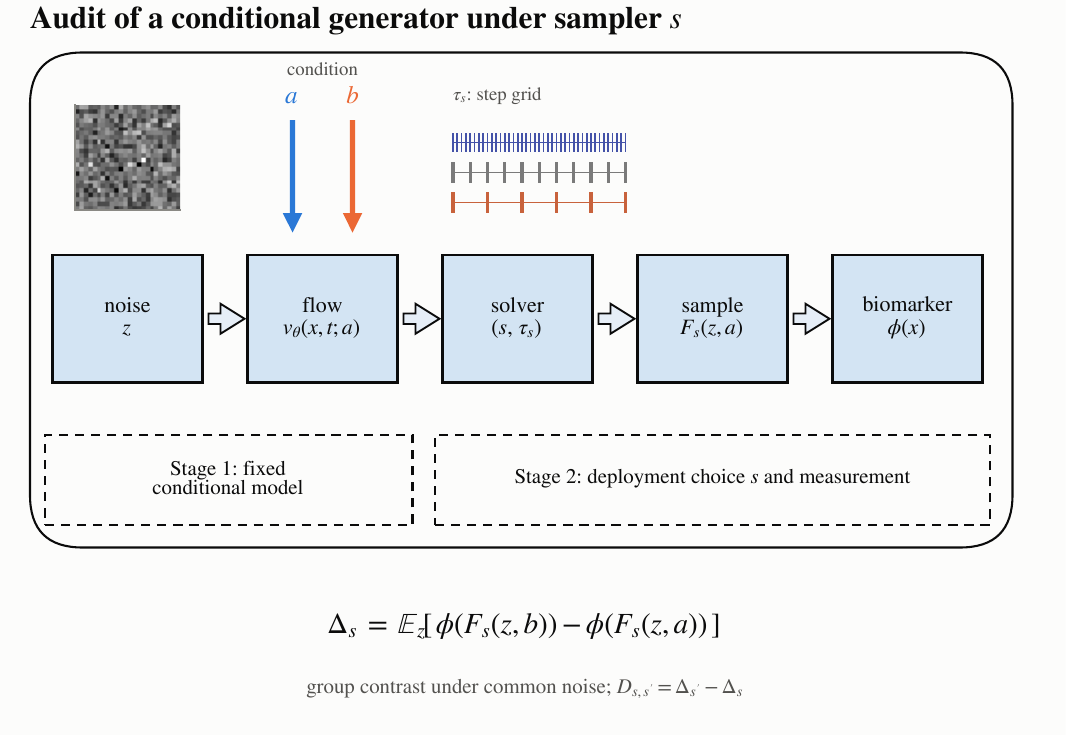}
\caption{\textbf{Sampling is part of the evaluated system.} With the model and random input held fixed, changing the solver or sampling steps may change the generated output. Any subgroup comparison therefore applies to the sampler tested. The diagram illustrates image-derived group contrasts, which are distinct from the clinical task-error gap $D(s)$ in Eq.~\eqref{eq:change}.}
\label{fig:policy}
\end{figure}

Figure~\ref{fig:policy} shows what stays fixed and what changes when samplers are compared. The model weights and starting noise can remain the same, while the resulting image and clinical measurements may differ. Evaluation should test the task error for each patient group under each relevant sampler using independent references. Intermediate states can help explain where clinically relevant errors arise.

\section{Why endpoint fairness may not transfer between samplers}

Fairness results obtained with one sampling procedure do not necessarily remain valid when that procedure changes. To understand why, consider two groups whose generation trajectories differ but lead to identical outputs under the exact continuous dynamics. A reference numerical sampler may also produce identical endpoints for both groups. However, changing the number of sampling steps can introduce numerical errors that differ between the groups, even though the trained model and its underlying dynamics remain unchanged.

\begin{proposition}[Fairness under one sampler need not hold under another]\label{prop:transfer}
For any reference step count $N\geq2$, any different positive integer $m$, and any $M>0$, there exist two smooth group-conditional ODEs whose exact solutions have identical endpoints and whose outputs also agree under $N$-step forward Euler sampling. When the same ODEs are evaluated using $m$-step forward Euler, the absolute difference between their numerical endpoints is at least $M$. The vector fields remain unchanged across the two sampling procedures. An analogous result holds for an additive-noise SDE when the groups are coupled using the same Brownian path.
\end{proposition}

The construction uses a quadratic time-dependent drift chosen so that its integral over the full generation interval is zero. Although the exact endpoints agree, forward Euler introduces a difference between the numerical solutions:
\begin{equation}\label{eq:gap}
\widehat x_b^{(k)}(1)-\widehat x_a^{(k)}(1)
=K\,\frac{N-k}{k^2(3N-1)},
\end{equation}
where $K>0$ determines the size of the difference. The numerical endpoints are identical when $k=N$, but they differ for any alternative step count $k=m$. If both groups have the same target and task error is measured using absolute endpoint loss, this numerical difference also produces unequal task errors. Appendix~\ref{supp:proof} provides the full derivation, an analytical illustration, and the corresponding stochastic argument. The result establishes that agreement under one sampler does not guarantee agreement under another.

A separate issue arises when another part of the system uses intermediate generation states. Two trajectories can begin and end at the same values while passing through different intermediate states. If a downstream model uses one of these states for editing, distillation, or another operation, the differences may influence its output even when the original generator produces identical final images (Appendix~\ref{supp:reuse}). When only the final output enters a clinical decision, the task error at the endpoint remains the appropriate outcome.

Measurements of generation paths must therefore be interpreted in the context of the sampling procedure and intended clinical task. Quantities such as trajectory energy and distance traveled can change with the time parameterization or numerical step grid, while stochastic comparisons also depend on how random noise is coupled across groups. The clinical significance of path measurements rests on their relationship to task error. Appendix~\ref{supp:bound} provides a bound on changes in task error and discusses its limitations for measurements such as vessel connectivity, where small image changes can produce substantial differences in the measured outcome.

\section{A clinical example for testing the position}

A useful clinical evaluation would examine a generative model whose outputs can be compared with independent reference measurements. Retinal-vessel segmentation or reconstruction provides one possible setting, particularly when the generated output is used to estimate vessel density, continuity, or other measures of vascular structure. A small vessel branch may be lost or incorrectly reconstructed without producing a large change in overall image similarity, yet the resulting error could affect a clinically relevant measurement. Previous work on retinal imaging and topology-aware segmentation provides methods for evaluating the preservation of these structures~\cite{brownPhysicsinformedDeepGenerative2024,shit2021cldice,gegundez2012evaluation,wu2024medsegdiff}. Research on retinal biomarkers and cardiovascular risk prediction further illustrates the importance of relating image-derived measurements to a clearly defined clinical outcome~\cite{raghavan2026retoncomace}.

Such a study could evaluate the same patient images using a fixed model checkpoint under a reference sampler and several faster alternatives. The resulting segmentations or reconstructed images would be compared with expert annotations or independent imaging references, allowing task errors to be measured for each patient. The analysis would then examine whether changing the sampling procedure produces larger increases in error for some patient groups than others, while accounting for clinically relevant differences in anatomy, disease severity, and acquisition conditions. This design would distinguish changes caused by numerical sampling from differences that may already exist between patient populations.

Our exploratory analysis with the age-conditioned WaveDiT flow model~\cite{danese2026wavedit} provides a simpler illustration of why sampling choices deserve attention. With the model parameters and initial random noise held fixed, reducing the number of sampling steps changed the contrasts between age conditions in two image-derived measurements. The absolute contrast increased for one measurement and decreased for the other, while changing the placement of sampling steps also affected agreement with the higher-step reference. Appendix~\ref{supp:illustration} presents the figures, numerical results, and reported uncertainty estimates. The opposing changes in these measurements show that the effect of sampling depends on which image features are used to assess the generated output.

A patient-level evaluation would need to determine whether changes in sampling produce clinically meaningful increases in task error and whether those increases differ across groups. Intermediate generation states could then be examined to identify when clinically relevant information is lost or altered and whether those observations help predict which patient cases are most affected by faster sampling. To establish their practical value, such measurements should be compared with a strong endpoint-only evaluation using the same patients, clinical references, and held-out testing procedure. The comparison would establish whether intermediate states provide useful information beyond the final outputs of the deployed sampler.

\section{Implications and limits}

Fairness evaluation should clearly specify the system being tested, including the patient population, model checkpoint, sampling procedure, postprocessing, clinical task, reference standard, and subgroup definitions. Results obtained under one sampling procedure should not be assumed to hold when that procedure changes. Previous studies have shown that the reliability of clinical AI depends on implementation, external validation, clinical workflows, and changes after deployment~\cite{wu2021fda,zhang2022deployment,who2021ethics,fda2025aidsf,azad2026readiness}. The sampling procedure should therefore be considered part of the system being evaluated.

When the number or timing of sampling steps changes, the model should be evaluated using the same patient cases and appropriate clinical reference measurements. Comparisons should account for randomness and report task errors for each subgroup with uncertainty estimates. Differences relative to a reference sampler can identify changes caused by numerical sampling, but they do not establish clinical accuracy. Examining intermediate states may help explain where these changes arise and which stages contribute to errors. However, their usefulness should be assessed against endpoint-only evaluation, with appropriate statistical controls.

A reasonable question is whether evaluating the final outputs of the deployed sampler is sufficient. For a fixed sampling procedure that has been properly validated, this may be the case. Our concern is whether those findings remain valid when the sampling procedure changes or intermediate states are used elsewhere. If clinical task errors remain stable across sampling procedures, or examining intermediate states provides no additional information, further trajectory analysis may not be necessary.

The mathematical construction rules out a general transfer guarantee; it does not estimate how frequently clinically important disparities arise. The exploratory MRI experiment uses 16 paired random seeds and two synthetic age inputs, without recruiting patients or using patient-level clinical data. Its image-derived measures are threshold-based proxies, and the 40-step sampler is a numerical reference rather than an independent clinical target. The bootstrap intervals and paired tests summarize variation over generated samples, not patients, and the significance markers in Figure~\ref{fig:transfer} are nominal and unadjusted for multiple comparisons. The study consequently establishes sensitivity of synthetic-image measurements to the sampler rather than patient-level clinical unfairness. The stochastic extension assumes shared additive noise, while the error bound requires a Lipschitz loss; these assumptions should be retained when applying the corresponding results.

\section{Conclusion}

Fairness evaluation for diffusion and flow-based models should consider the sampling procedure used to generate images and the clinical task those images are intended to support. Previous research has shown that sampling choices affect generation and that clinically relevant information may change throughout the process. Our mathematical example shows that equal outputs under one sampler do not guarantee equal outputs under another. These findings support evaluating clinical task errors across relevant sampling procedures and patient groups, with intermediate states examined when they help explain observed differences. Such evaluation can help determine whether changes in sampling affect the reliability of clinical measurements and whether those effects differ across patient groups.

\paragraph{Code and data availability.}
The analysis code is publicly available at \url{https://github.com/rdharini2001/generation-path-fairness}.

\clearpage
\appendix
\section*{Supplementary material}
\addcontentsline{toc}{section}{Supplementary material}
\noindent This supplement gives the mathematical derivations for the sampler-transfer result and the detailed numerical comparisons from the age-conditioned MRI synthesis experiment.

\section{Age-conditioned sampling: an exploratory example}\label{supp:illustration}

We examine the WaveDiT image flow model under two age conditions, 30 and 80 years, using 16 matched random seeds~\cite{danese2026wavedit}. The trained model and initial noise remain fixed as we compare a five-step sampler with a 40-step numerical reference. We also compare two ten-step schedules that place more steps earlier or later in generation. The outputs are assessed using two image-derived quantities, Biomarker A and Biomarker B. Biomarker A is the fraction of masked voxels in the lowest intensity class of a three-class Otsu partition. Biomarker B is the foreground-mask voxel count; its between-condition contrast is normalized by each seed's mean foreground count under the 40-step reference. Both measures are computed from synthetic images using the same image-processing rules across samplers.

Let $m_r$ denote the measurement of biomarker $r$. The difference between age conditions $a$ and $b$ under sampling procedure $\pi$ is
\begin{equation}\label{eq:contrast}
 C_r(\pi)=\E_Z\bigl[m_r(G_{\theta,\pi}(Z,b))-m_r(G_{\theta,\pi}(Z,a))\bigr],
 \qquad \Delta C_r=C_r(\pi)-C_r(\pi_0).
\end{equation}
Here, $\pi_0$ is the reference sampler. The contrast describes how a measurement differs between the two generated age conditions. We distinguish the change in signed contrast, $\Delta C_r$, from the change in its magnitude, $|C_r|$.

\begin{table}[h]
\centering\small
\caption{Age-condition contrasts under 40-step and five-step sampling. The first two numerical columns give contrast magnitudes; the change, confidence interval, and $p$-value refer to the signed contrast across matched seeds.}
\label{tab:reported}
\begin{tabular}{@{}lrrrll@{}}\toprule
Quantity & $|C_r(40)|$ & $|C_r(5)|$ & $\Delta C_r$ & 95\% CI & $p$\\\midrule
Biomarker A & 0.0201 & 0.0328 & +0.0126 & [0.0066, 0.0193] & 0.0007\\
Biomarker B & 0.0778 & 0.0484 & +0.0294 & [0.0213, 0.0375] & 0.0002\\\bottomrule
\end{tabular}
\end{table}

Reducing sampling from 40 to five steps increased the absolute contrast for Biomarker A by approximately 63\%, while the absolute contrast for Biomarker B decreased by approximately 38\%. The signed changes were $+0.0126$ for A and $+0.0294$ for B (Table~\ref{tab:reported}). In Biomarker B, the positive signed change reflects a reduction in the magnitude of an initially negative contrast. The slight difference between the rounded values shown for Biomarker A and its signed change arises from rounding.

At a fixed budget of ten steps, placing more steps earlier in generation gave a smaller mean absolute deviation in Biomarker A from the 40-step reference than placing more steps later. The reported difference was 0.0079 (95\% CI [0.0049, 0.0112]; $p=0.0001$). The step schedule thus changes how closely the faster sampler reproduces the 40-step output.

Together, these comparisons show that sampling budget and step placement affect distinct image-derived contrasts in different ways. The underlying synthetic measurements and analysis code are available at \url{https://github.com/rdharini2001/generation-path-fairness}.

\subsection*{Further sampling comparisons}
\begin{figure}[!htbp]
\centering
\includegraphics[width=0.96\linewidth]{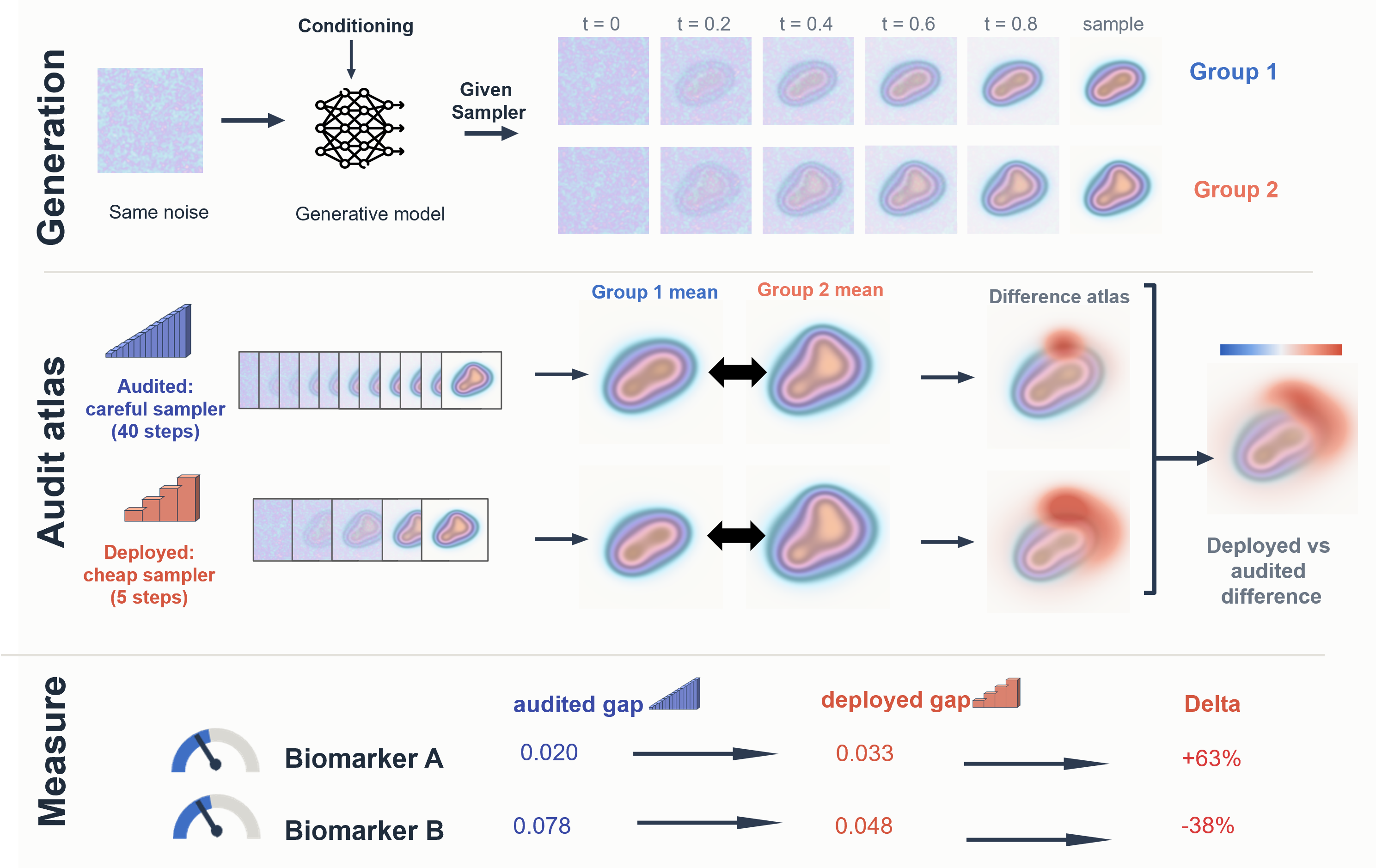}
\caption{\textbf{Age-conditioned measurements under two samplers.} The generator uses the same trained weights and matched initial noise under a 40-step reference and a five-step sampler. The two image-derived group contrasts change in different directions. The images are generated under the two age conditions.}
\label{fig:example}
\end{figure}

\begin{figure}[!htbp]
\centering
\includegraphics[width=0.96\linewidth]{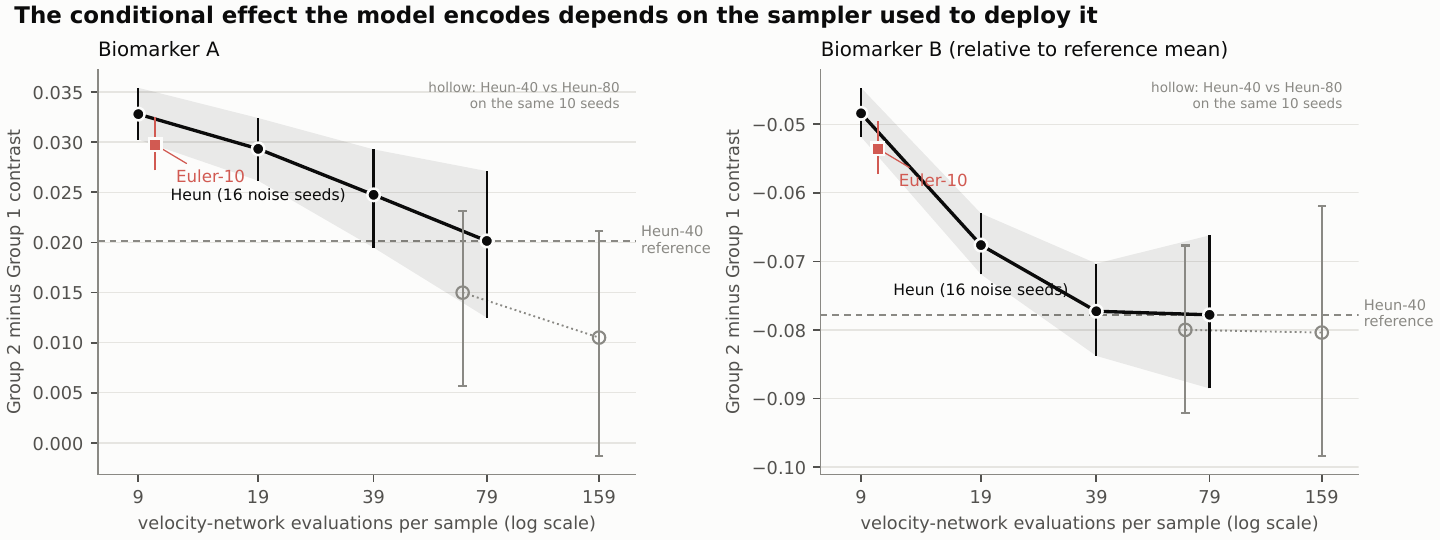}
\caption{\textbf{Biomarker contrasts across sampling budgets.} Biomarker A and Biomarker B respond differently as the number of steps changes. Bands and error bars summarize variation across matched synthetic samples.}
\label{fig:cost}
\end{figure}

\begin{figure}[!htbp]
\centering
\includegraphics[width=0.98\linewidth]{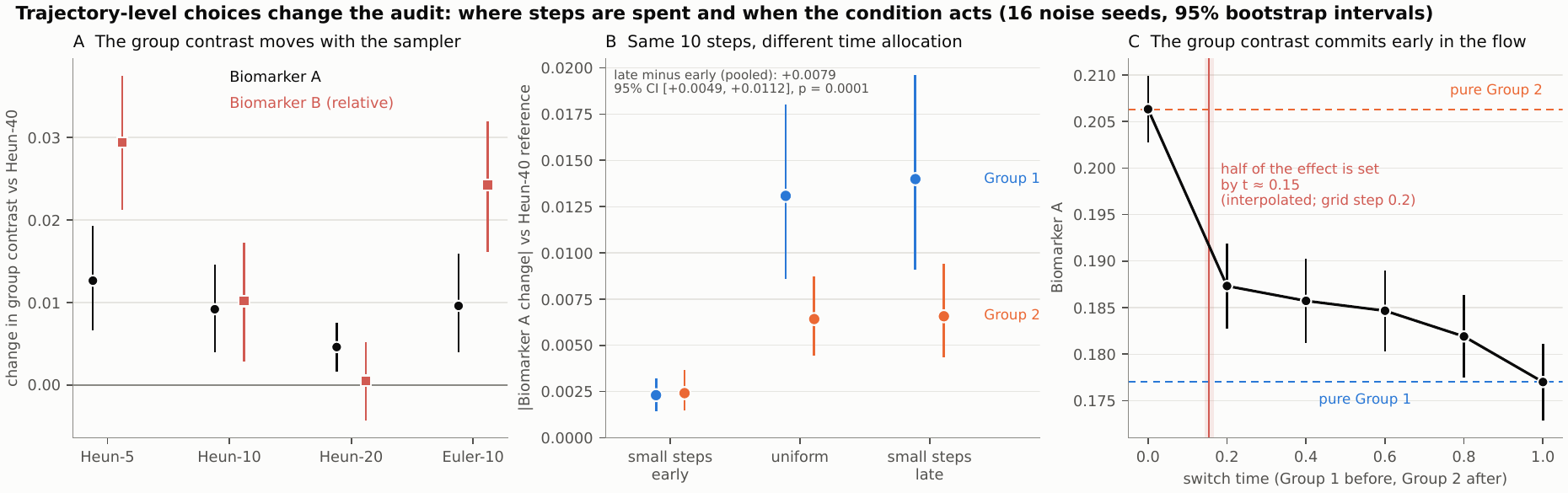}
\caption{\textbf{Changes in the sampling procedure.} The panels compare solvers and step schedules (left), early and late allocation of ten sampling steps (middle), and the timing of an introduced conditional effect (right). Error bars summarize variation over paired synthetic samples; the final panel illustrates the response to a change in conditioning time.}
\label{fig:time}
\end{figure}

\begin{figure}[!htbp]
\centering
\includegraphics[width=0.96\linewidth]{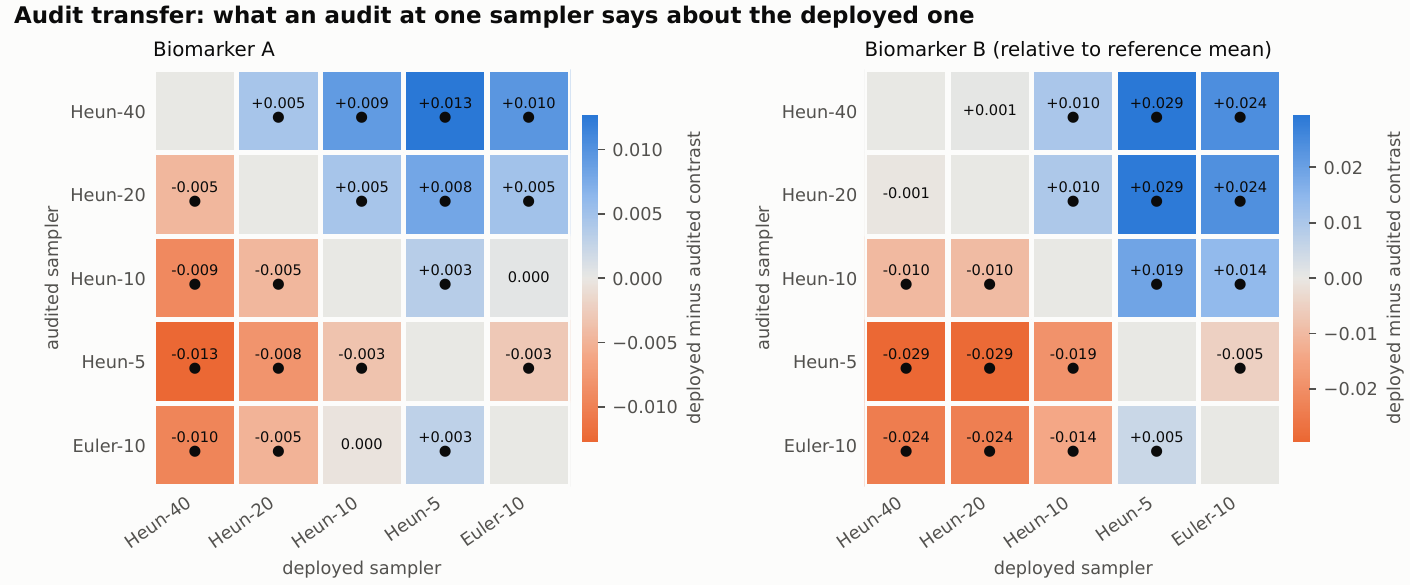}
\caption{\textbf{Group contrasts under different sampler comparisons.} Each cell shows the change in an image-derived group contrast between a candidate sampler and a chosen reference. The heat maps show how the measured contrast depends on the reference and candidate samplers. Significance markers denote nominal tests, unadjusted for multiple comparisons.}
\label{fig:transfer}
\end{figure}

Figure~\ref{fig:example} shows the two age-conditioned comparisons. Figure~\ref{fig:cost} shows how the measured contrasts vary with the number of sampling steps. Figure~\ref{fig:time} compares step placement and other changes to the generation procedure. Figure~\ref{fig:transfer} compares contrasts obtained under different reference and candidate samplers. All four figures use the same synthetic-image setting and measurement definitions.

\clearpage
\section{A mathematical counterexample to fairness transfer between samplers}\label{supp:proof}

Fix $N\ge2$ and consider two smooth scalar ODEs on $0\le t\le1$:
\begin{equation}\label{eq:toy}
 \dot x_g(t)=gK h_N(t),\qquad x_g(0)=0,\qquad
 h_N(t)=1-2t+\frac{6N}{3N-1}\left(t^2-\frac13\right),\quad g\in\{0,1\}.
\end{equation}
The drift does not depend on $x$, and $K>0$ controls its scale. Since $\int_0^1h_N(t)\,\dd t=0$, both groups have exact endpoint zero. With $k$ uniform forward Euler steps, the difference between the numerical endpoints is
\begin{align}\label{eq:sums}
\widehat x_1^{\,(k)}(1)-\widehat x_0^{\,(k)}(1)
 &=\frac Kk\sum_{j=0}^{k-1}h_N(j/k),\\
\frac1k\sum_{j=0}^{k-1}\left(1-\frac{2j}{k}\right)&=\frac1k,\qquad
\frac1k\sum_{j=0}^{k-1}\left(\frac{j^2}{k^2}-\frac13\right)=-\frac1{2k}+\frac1{6k^2}.\nonumber
\end{align}
Substituting these sums gives
\begin{equation}\label{eq:counterformula}
 \widehat x_1^{\,(k)}(1)-\widehat x_0^{\,(k)}(1)=K\frac{N-k}{k^2(3N-1)}.
\end{equation}
The endpoints agree when $k=N$. For any $m\ne N$, the difference is nonzero. Choosing $K\ge M m^2(3N-1)/|N-m|$ makes its magnitude at least $M$. Thus, equal exact endpoints and equal outputs under the $N$-step reference sampler do not guarantee equal outputs under another step count. This proves Proposition~\ref{prop:transfer} for the ODE construction. If both groups have target $y^*=0$ and loss $|\widehat x_g-y^*|$, group 0 has zero error, whereas group 1 has the absolute error in Eq.~\eqref{eq:counterformula} under the changed sampler. The corresponding group errors therefore agree under the reference sampler but differ under the alternative discretization.

\begin{table}[h]
\centering\small
\caption{Exact endpoint differences for $N=40$ and $K=1$. The groups agree under the continuous-time solution and the 40-step sampler, but not under the other step counts. All values follow from Eq.~\eqref{eq:counterformula}.}
\label{tab:toy}
\begin{tabular}{@{}lrr@{}}\toprule
Euler steps $k$ & Signed difference & Absolute difference\\\midrule
5 & $+0.0117647$ & $0.0117647$\\
10 & $+0.0025210$ & $0.0025210$\\
20 & $+0.0004202$ & $0.0004202$\\
40 (reference) & $0$ & $0$\\
80 & $-0.0000525$ & $0.0000525$\\\bottomrule
\end{tabular}
\end{table}

\begin{figure}[h]
\centering
\includegraphics[width=.84\linewidth]{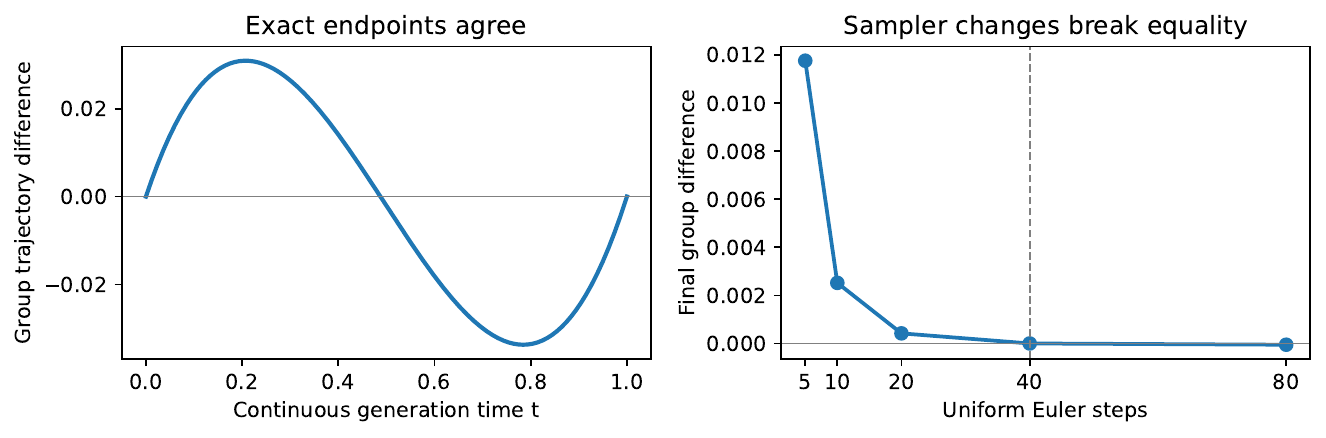}
\caption{\textbf{Exact trajectories and numerical endpoints.} The difference between the two exact trajectories returns to zero at the endpoint (left). Forward Euler gives a nonzero endpoint difference when the step count differs from the 40-step reference (right). The curves represent the exact analytical construction.}
\label{fig:analytic}
\end{figure}

\subsection*{Extension to additive stochastic noise}
Consider $\dd X_t^g=gK h_N(t)\,\dd t+\sigma\,\dd W_t$, with identical initial conditions and diffusion coefficient for both groups. We use the same Brownian path for the two groups and across sampling grids. The exact drift difference integrates to zero at $t=1$, while the shared noise cancels when the group endpoints are subtracted. Under Euler--Maruyama, the Brownian increments on each grid sum to the same $\sigma W_1$. The group difference is therefore still given by Eq.~\eqref{eq:counterformula}. This argument relies on shared additive noise. It does not give the same expression for independent noise, different group diffusion coefficients, or state-dependent diffusion. The choice of coupling must be stated when interpreting a stochastic comparison.

\section{Why using intermediate states changes the system}\label{supp:reuse}

Consider $\dot x_g(t)=gK(1-2t)$ with $x_g(0)=0$. Its solution is $x_g(t)=gKt(1-t)$, so both groups have endpoint zero. At an intermediate time $\tau\in(0,1)$, however, their states differ by $K\tau(1-\tau)$, which is $K/4$ when $\tau=1/2$. A model that uses $x_\tau$ for editing, distillation, or another task may therefore respond differently across groups even though the original final outputs agree. This explains why intermediate states can matter when they are used by another part of the system. The relevance of this difference follows from how the downstream method uses the intermediate state.

\section{A bound on changes in task error}\label{supp:bound}

Suppose $f_y(x)=\ell(T(x),y)$ is $L_f$-Lipschitz in $x$ for the relevant clinical references $y$. Couple the outputs of two samplers for the same patient and random input. The change in standardized task risk for group $g$ is then bounded by the expected difference between their outputs:
\[
 |R_g(\pi)-R_g(\pi_0)|\le L_f\int \E[\|G_{\theta,\pi}(U,Z)-G_{\theta,\pi_0}(U,Z)\|\mid A=g,Q=q]w(q)\,\dd q.
\]
Applying the triangle inequality to the two groups gives
\begin{equation}\label{eq:taskbound}
 |D(\pi)-D(\pi_0)|\le L_f\sum_{g\in\{a,b\}}\int \E[\|G_{\theta,\pi}-G_{\theta,\pi_0}\|\mid A=g,Q=q]w(q)\,\dd q.
\end{equation}
This bounds the change in the subgroup task-error gap, not the gap itself. It also depends on the assumed smoothness of the loss. A small change in an image can produce a large change in a thresholded diagnosis, vessel connectivity, or topology, so the bound may not apply to those measurements.

A related bound holds for stochastic dynamics. Let $\dd X_t^g=b_g(t,X_t^g)\,\dd t+\sigma_t\,\dd W_t$, with identical initial conditions and shared additive noise. If $b_a(t,\cdot)$ is $L_b$-Lipschitz, then synchronous coupling gives
\[
 \E\|X_1^b-X_1^a\|\le \int_0^1 e^{L_b(1-s)}\E\|b_b(s,X_s^b)-b_a(s,X_s^b)\|\,\dd s.
\]
To obtain this inequality, subtract the two equations, add and subtract $b_a(s,X_s^b)$, and apply Gronwall's inequality. The Wasserstein-1 distance between the endpoint distributions is at most the expected distance under this coupling. The inequality gives a stability guarantee under the stated coupling and regularity assumptions. Path length and energy also depend on the choice of time scale, so their differences cannot be interpreted as unfairness on their own.

\section{How to test the position with clinical data}\label{supp:design}

A clinical evaluation should compare realistic sampling procedures on the same patients using independent task references, such as expert-labelled vessel segmentations or images obtained from an independent acquisition. The model checkpoint, patient inputs, and conditioning would remain fixed while the solver, step count, or step schedule changes. Randomness should be paired where the sampling methods permit it. For each subgroup, the study should report absolute task errors and uncertainty, then examine changes in those errors after adjustment for clinically justified variables. The samplers, outcomes, and acceptable error differences should be chosen before the analysis. For vascular tasks, useful outcomes might include segmentation error, vessel continuity, and deviation in vessel density. Each must have a clear clinical purpose, and thresholds or model settings should be selected without using the held-out test cases.

The additional value of intermediate states should be tested directly. One model could predict which patients have a clinically meaningful increase in error after acceleration using only endpoint information; a second could use the same information together with trajectory measurements. Both models should be trained and tuned on training patients, then evaluated on held-out patients with comparable access to data and model capacity. If trajectory measurements do not improve prediction, they are unlikely to justify routine use. If they do help, the findings should be tested across imaging devices, clinical sites, and patient populations. Stable subgroup errors across the samplers studied would support endpoint-only evaluation for that setting, even though the mathematical counterexample rules out a general guarantee across all samplers.

\end{document}